\documentclass[11pt,a4paper]{article}
\usepackage[hyperref]{emnlp2020}
\usepackage{times}
\usepackage{latexsym}
\usepackage{graphicx}
\usepackage{epstopdf}
\usepackage{amssymb}
\usepackage{amsmath}
\usepackage{multirow}
\usepackage{xspace}
\usepackage{enumitem}

\usepackage{microtype}

\aclfinalcopy 

\newcommand{\our}{\mbox{MULTICAST}\xspace}

\newcommand{\ie}{\textit{i.e.}}

\title{Empower Distantly Supervised Relation Extraction with \\Collaborative Adversarial Training}

\author {
    Tao Chen\textsuperscript{\rm 1},
    Haochen Shi\textsuperscript{\rm 1},
    Liyuan Liu\textsuperscript{\rm 2},
    Siliang Tang\textsuperscript{\rm 1}\thanks{\quad Corresponding author} ,
    Jian Shao\textsuperscript{\rm 1}, \\
    \textbf{Zhigang Chen}\textsuperscript{\rm 3} \textbf{\&}
    \textbf{Yueting Zhuang}\textsuperscript{\rm 1}
    \\
    \textsuperscript{\rm 1}Zhejiang University
    \textsuperscript{\rm 2}University of Illinois at Urbana Champaign
    \textsuperscript{\rm 3}iFLYTEK Research \\
    
    \texttt{\{ttc, hcshi, siliang, jshao,  yzhuang\}@zju.edu.cn} \\
    \texttt{llychinalz@gmail.com, zgchen@iflytek.com}
}

\date{}

\begin{document}
\maketitle

\begin{abstract}
With recent advances in distantly supervised (DS) relation extraction (RE), 
considerable attention is attracted to leverage multi-instance learning (MIL) to distill high-quality supervision from the noisy DS. 
Here, we go beyond label noise and identify the key bottleneck of DS-MIL to be its \emph{low data utilization}: 
as high-quality supervision being refined by MIL, MIL abandons a large amount of training instances, which leads to a low data utilization and hinders model training from having abundant supervision. 
In this paper, we propose collaborative adversarial training to improve the data utilization, which coordinates virtual adversarial training (VAT) and adversarial training (AT) at different levels.
Specifically, since VAT is label-free, we employ the instance-level VAT to recycle instances abandoned by MIL.
Besides, we deploy AT at the bag-level to unleash the full potential of the high-quality supervision got by MIL. 
Our proposed method brings consistent improvements ($\sim 5$ absolute AUC score) to the previous state of the art, which verifies the importance of the data utilization issue and the effectiveness of our method. 
\end{abstract}

\section{Introduction}
\label{sec:intro}

Relation extraction (RE) aims at identifying the relation between entities within a specific context and provides essential support for many downstream tasks. 
As the performance of RE systems is generally limited by the amount of training data, recent RE systems typically resort to distant supervision (DS) to fetch abundant training data by aligning knowledge bases (KBs) and texts.
Since this strategy inevitably introduces label noise to model training, how to neutralize the label noise has been viewed as a major problem of DS.  

Multi-instance learning (MIL) was introduced to handle label noise~\citep{zeng2015distant,lin2016neural} and has received a significant amount of attention. 
Specifically, MIL clusters training instances into bags.
For each bag, MIL demotes its low-quality instances to eliminate label noise and refines high-quality instances as the bag-level representation for model training. 

\begin{figure}[t]
\centerline{\includegraphics[width=\linewidth]{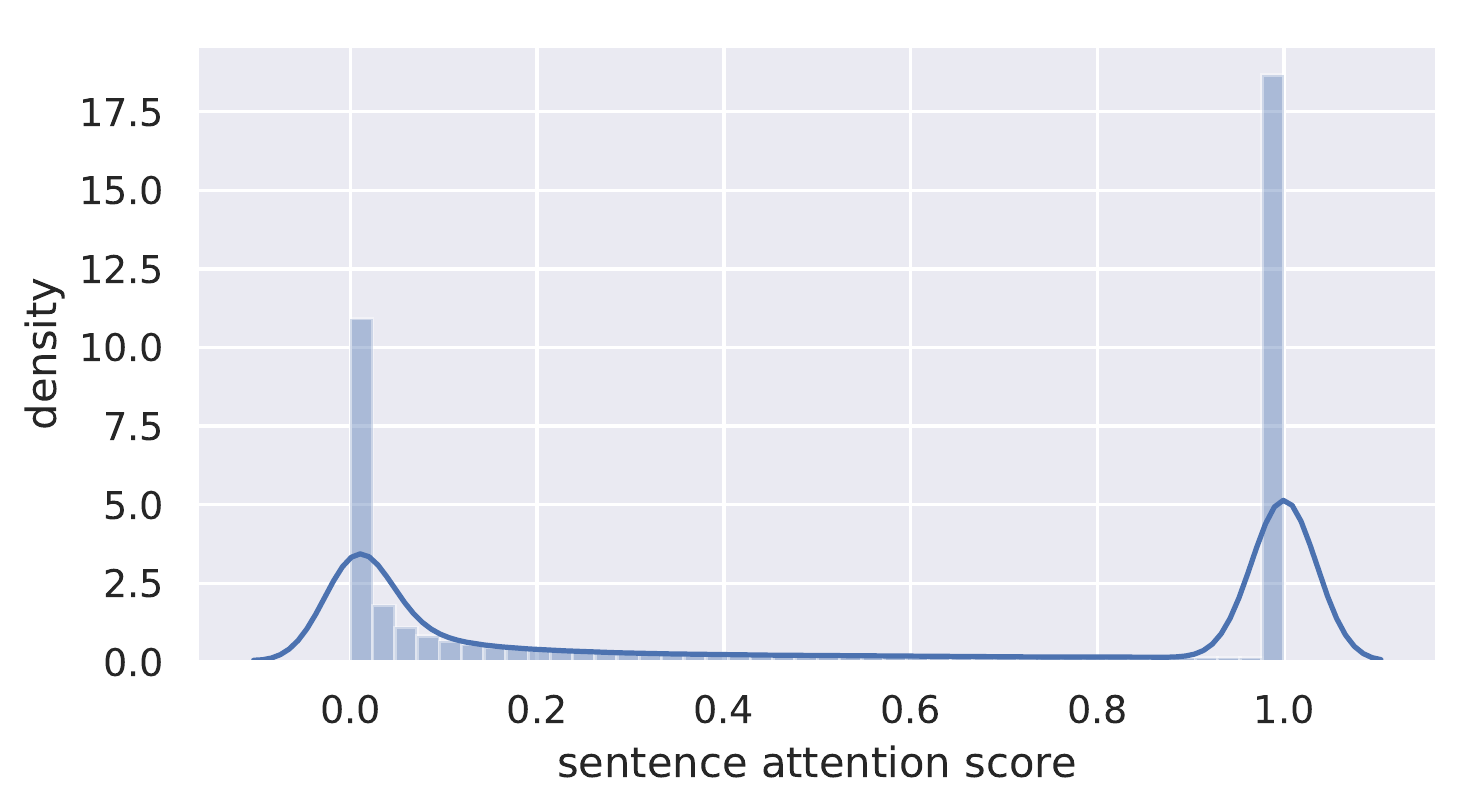}}
\caption{Sentence attention score distribution inside the bag during training process: Most instances are with low scores, instances with high attention scores (excluding 1.0) only dominate a small part of data.}
\label{sentence bag}
\end{figure}

Here, we go beyond label noise and identify the key bottleneck of DS-MIL to be its low data utilization. 
In order to distill high-quality supervision from DS, MIL only focuses on a few representative instances (with high attention scores) and abandons a large proportion of low-score instances. 
As in Figure~\ref{sentence bag}, except the situation that one bag only contains one instance (with attention scores of 1.0), most of the instances are assigned with low attention scores (0.0$\sim$0.2) and abandoned during the training process.
Specifically, as in Table~\ref{control exp}, control experiments show that even if some low-score instances are removed, the newly trained model has a limited performance change.
In other words, although DS leads to abundant training instances, MIL fails to unleash the full potential of DS, since it abandons the majority of training instances. 

Here, we propose \our (\textbf{MULT}i-\textbf{I}nstance \textbf{C}ollaborative \textbf{A}dver\textbf{S}arial \textbf{T}raining) to improve the data utilization. 
It coordinates adversarial training (AT)~\citep{goodfellow2014explaining} and virtual adversarial training (VAT)~\citep{ miyato2016adversarial} at different levels. 
In detail, as the MIL framework intrinsically splits training data into two classes (i.e., high-quality instances for constructing bag-level representations and low-quality instances abandoned by MIL), we use different strategies on them.
For low-quality instances, although their associated labels are not very reliable, they can still provide valuable information for label-free regularization objectives. 
Thus, we apply instance-level virtual adversarial training (IVAT) to exploit entity and context information without using their unreliable label information.
For high-quality instances, we try to compensate their loss of quantity (caused by MIL). 
Specifically, we apply bag-level adversarial training (BAT) to further regularize the constructed representations and unleash the full potential of these high-quality instances. 

We conduct experiments on NYT~\citep{riedel2010modeling}, the public DSRE benchmark.
\our leads to consistent improvements over the previous state-of-the-art systems. 
It demonstrates the effectiveness of \our and validates our intuition that the data utilization issue is the key bottleneck. 
We further conduct ablation studies to verify that \our coordinates different modules effectively. 
The major contributions of this paper are summarized as follows:
\begin{itemize}[leftmargin=*,nosep]
    \item We identify the low data utilization issue as the major bottleneck of DS-MIL.
    \item We propose \our to boost data utilization. It coordinates VAT and AT at different levels based on MIL signals (attention scores). 
    \item Controlled experiments verify our intuitions and show that \our leads to consistent improvements ($\sim 5$ absolute AUC score). 
\end{itemize}

\begin{figure*}
\centerline{\includegraphics[width=\linewidth]{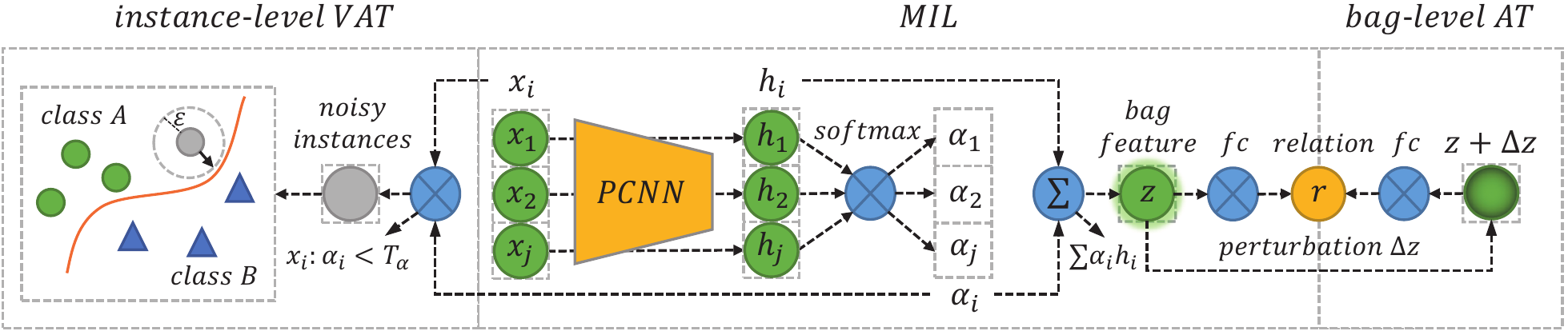}}
\caption{(a) Instances $x_1,x_2\dots x_j$ inside the bag firstly encode themselves by piecewise convolutional neural networks, and obtain sentence-level representations $h_1,h_2\dots h_j$. Based on the MIL framework, selective attention is adopted to form bag-level representations $z=\sum_i \alpha_ih_i$ over instances. (b) Inside the bag, those noisy or unrepresentative instances with lower attention score $\alpha_i$ are selected $\{x_i|\alpha_i<T_\alpha\}$ for additional virtual adversarial training. (c) Outside the bag, reliable bag-level representations $z$ are further enhanced via adversarial learning.}
\label{Model Framework}
\end{figure*}

\section{Related Work}
\label{sec:related work}

In the field of distantly supervised relation extraction, the multi-instance learning framework is introduced to handle the label noise of DS.
Recently, MIL has become a common paradigm for DSRE and many efforts have been made for further improvements~\citep{lin2016neural, qin2018dsgan, ye2019distant, yuan2019cross, huang2019self, ye2019looking, shang2020noisy}.
In these MIL frameworks, sentences are first encoded by handcrafted features~\citep{mintz2009distant, hoffmann2011knowledge} or neural networks. 
Then, multiple instances are leveraged to form a bag-level representation, which has better quality. 
With regard to the strategy for selecting instances inside the bag, a soft attention mechanism~\citep{lin2016neural} is widely used for its better performance than the hard selection way.

Based on the multi-instance learning framework, most previous work focus on further improving the strategy to handle label noise.
Specifically, ~\citet{ye2019distant, yuan2019cross} both adopted a relation-aware selective attention mechanism inside the bag, and constructed a superbag which contains a group of bags to alleviate the issue of bag label error. Focusing on transforming the network structure, ~\citet{huang2019self} utilized recent self-attention mechanism~\citep{vaswani2017attention} integrated with convolutional neural networks (CNNs) to obtain a better sentence representation from the noisy inputs, and this work also applied cooperative curriculum learning to constrain student models which can learn from each other. 
At the same time, few attempts have been made on other aspects of DSRE, i.e., ~\citet{ye2019looking} found that the problem of shifted label distribution influences the performance of DSRE models significantly.
Similar to our study, \citet{shang2020noisy} observe that noisy sentences inside the bag are not useless and developed a way to relabel the noisy data by employing unsupervised deep clustering.

At the same time, adversarial training has been found to be useful for DSRE. 
~\citet{wu2017adversarial} firstly introduced adversarial training~\citep{goodfellow2014explaining, miyato2016adversarial} to relation extraction by generating adversarial noise to the training data. 
\citet{qin2018dsgan} leverages generative adversarial networks (GANs), i.e., it adopts the trained generator to filter the DS training dataset and redistributes the false positive instances into the negative set, in which way to provide a cleaned dataset for relation classification.

\section{Multi-Instance Collaborative Adversarial Training}
\label{sec:methodology}

In our paper, we identify the low data utilization as the key bottleneck of DS-MIL. 
As MIL forms accurate bag representations to handle label noise, it abandons a large amount of training instances. 
Typically, MIL faces the dilemma that label noise reduction sacrifices the data utilization. 
Here, we go beyond typical DS-MIL and propose collaborative adversarial training to improve the data utilization. 
The diagram of our method (\our) is visualized in Figure~\ref{Model Framework}, which contains five components: (1) input representations; (2) sentence encoder; (3) attention-based MIL framework; (4) instance-level virtual adversarial training module; (5) bag-level adversarial training module. 

\subsection{Inputs: Embeddings}

For each word $t_i$ in sentence $s$, we employ word embedding $w_i\subset \mathbb{R}^{d_w}$ to capture its semantic information.
Moreover, to encode the sentence in an entity-aware manner, relative position embedding~\citep{zeng2015distant} is leveraged to represent the position information in the sentence. 
Relative distances $d_{i1},d_{i2}$ of word $t_i$ correspond to the distances between $t_i$ and two entities $e_1$ and $e_2$, and can be transferred to position vectors $p_{i1}, p_{i2} \subset \mathbb{R}^{d_p} $ by looking up a position embedding table. This embedding table is initialized randomly and updated during the training process.

\begin{figure}[htbp]
\centerline{\includegraphics[scale=0.8]{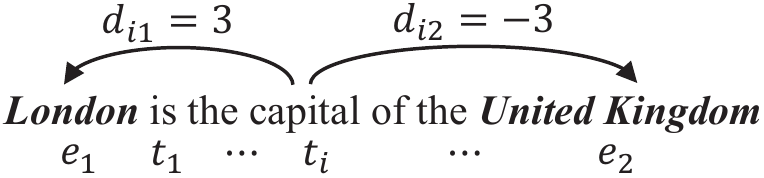}}
\caption{Relative position distance.}
\label{position distance}
\end{figure}

Concatenating the above two embeddings, each word $t_i$ can obtain its entity-aware representation as $m_i=[w_i;p_{i1};p_{i2}] \subset \mathbb{R}^{d}$. 
Thus the instance representation can be constructed as $X=[m_1;m_2;\dots;m_l] \subset \mathbb{R}^{l\times d}$, where $d = d_w + 2 \cdot d_p$ and $l$ is the maximum length of the sentences.

\subsection{Encoder: Piecewise CNN}

Convolutional neural networks capture the sentence semantics with sliding windows. 
In the convolutional layer, the embedding window $X_{t;t+u}=[m_t;m_{t+1}\dots;m_{t+u-1}] \subset \mathbb{R}^{u\times d}$ interacts with convolution kernels $\{W_1,\dots,W_p\} \subset \mathbb{R}^{u\times d}$ to extract sentence-level features, where $u$ is the width of kernel and $p$ is the number of kernels.

Followed by max-pooling layer, the most responsive region of convolutional output $C\subset \mathbb{R}^{l\times p}$ is retained. Instead of just using a unified pooling layer,~\citet{zeng2015distant} applied max-pooling operation to different pieces of sentence respectively, which has been proved to better capture structured information between two entities. The final feature vector $H\subset \mathbb{R}^{3\times p}$ can be obtained by concatenating all pooling results of three pieces. 

\subsection{MIL: Multi-Instance Learning}
\label{sec:selective attention}

For a model parameterized by $\theta$, input representation $x_i \in X$ of each sentence $s_i$ in bag $B$ can be encoded to feature vector $h_i \in H$, then multi-instance learning framework considers all instances inside the bag to get a relatively accurate representation $z$, which is defined as: \begin{equation*}
    z=\sum_i \alpha_i h_i
\end{equation*}

As for the weight $\alpha$, we adopt a soft attention mechanism as in~\citet{lin2016neural}, where $\alpha_i$ is the normalized attention score calculated by a query-based function $f_i$ which measures how well the sentence representation $h_i$ and the predict relation $r$ matches:
\begin{equation*}
    \alpha_i=\frac{e^{f_i}}{\sum_j e^{f_j}}
\end{equation*}
where $f_i=h_iA q_r$, $A$ is a weighted diagonal matrix and $q_r$ is
the query vector which indicates the representation of relation $r$ (randomly initialized).

Then, based on this bag-level representation, a simple fully-connected layer with activation function $\textit{softmax}$ is added to map the feature vector $z$ to a conditional probability distribution $p(r|Z,\theta)$:
\begin{equation*}
    p(r|Z,\theta)=\frac{e^{o_r}}{\sum_{i=1}^{n_r} e^{o_i}}
\end{equation*}
where $o=Mz+b$ is the score associated to all relation types, $n_r$ is the total number of relations, $M$ is a projection matrix and $b$ is the bias term.

Finally, we define the objective function of MIL framework using cross-entropy as follows:
\begin{equation*}
    J(\theta) = -\sum_{i=1} \log p(r_i|z_i,\theta)
\end{equation*}

\subsection{IVAT: Instance-Level Virtual Adversarial Training}
\label{sec:IVAT}

In MIL, normalized attention score $\alpha_i$ describes how much the instance $x_i$ contribute to the final representation $z$. 
A higher value indicates the instance is cleaner or more representative, while a lower value implies the instance is noisy (i.e., its relation label is not reliable). 
In other words, attention score is the label quality signal used in MIL. 

We refer instances with high attention scores as $X_{clean}$ (i.e., clean instances) and instances with low attention scores as $X_{noisy}$ (noisy instances). 
As discussed in Section~\ref{sec:intro}, MIL mainly focuses on $X_{clean}$ and abandons $X_{noisy}$ during the training. 
To improve the data utilization of MIL, we introduce virtual adversarial training at the instance level to exploit entity and context information from $X_{noisy}$. 
Now we proceed to introduce module details.

For instances $\left\{x_1,x_2,\dots,x_i\right\}$ inside bag $B$, we use $\left\{ \alpha_1, \alpha_2, \dots, \alpha_i \right\}$ to refer their normalized attention scores (outputs of the selective attention in section MIL). Then, we leverage a hyperparameter $T_\alpha$ to identify instances that are ignored by MIL:
\begin{equation*}
X_{noisy}=\{x_i|\alpha_i < T_\alpha \}
\end{equation*}

For instance $x \in X_{noisy}$, we refer its conditional probability distribution output to be $p(y|x, \theta)$. 
Then, we refer its representation under a small perturbation $||d||\leq \epsilon_x$ to be $x + d$, and the corresponding model output 
to be $p\left(y | x+d, \theta\right)$. 
These two outputs are regularized to be similar, i.e., 
\begin{equation*}
l_{\mathrm{ivat}}\left(d, x, \theta\right) := \mathrm{KL}\left[ p\left(y | x, \theta\right) \| p\left(y | x+d, \theta\right)\right]
\end{equation*}
where $\mathrm{KL}$ is the Kullback–Leibler divergence which measures the similarity of two probability distributions. 
As to the adversarial perturbation $d_{v-adv}$, its ideal choice should be the direction maximizing $l_{\mathrm{ivat}}$, i.e.,  
\begin{equation*}
    d_{v-adv} := \arg \max _{d}\left\{l_{\mathrm{ivat}}\left(d, x, \theta\right) ;\|d\|_{2} \leq \epsilon_x\right\}
\end{equation*}
Following previous work~\citep{miyato2018virtual}, we employ an efficient way to estimate $d_{v-adv}$ under $L_2$ norm:
\begin{equation*}
    d_{v-adv} \approx \epsilon_x \frac{g}{\|g\|_{2}}
\end{equation*}
where $g=\left.\nabla_{r} \mathrm{KL}\left[p(y | x, \theta), p(y | x+r, \theta)\right]\right|_{r=\xi d}$ with $\xi > 0$ and $d$ is a randomly sampled unit vector.
For neural networks, this approximation can be performed with $K$ sets of back-propagations.

With such a perturbation $d_{v-adv}$, our objective is to make the local distributional smoothness (LDS) of the model as high as possible, this is defined as:
\begin{equation*}
\operatorname{LDS-X}\left(\theta\right) := -\sum_{x \in X_{noisy}} l_{\mathrm{ivat}}\left(d_{v-adv}, x, \theta\right)
\end{equation*}

\subsection{BAT: Bag-Level Adversarial Training}
\label{sec:bat}

Different from noisy instances, high-quality instances are used to construct the bag-level representation $z$, which better matches the associated relation and allows MIL to reduce the impact of label noise. 
Here we leverage adversarial training to unlease the full potential of that high-quality supervision. 

Specifically, we add a perturbation $d$ to the bag-level representation $z$ instead of word embedding $x$. 
Different from IVAT, we employ the training label instead of the original output to regularize the output under perturbation, \ie, 
\begin{equation*}
    l_{\mathrm{bat}}\left(d, z, \theta\right) := - \log  p\left(r | z+d, \theta\right)
\end{equation*}

Similar to the virtual adversarial perturbation $d_{v-adv}$ in section IVAT, adversarial perturbation $d_{adv}$ is in the direction with maximum model output change, which is further defined as:
\begin{equation*}
    d_{adv} := \arg \max _{d}\left\{l_{\mathrm{bat}}\left(d, z, \theta\right) ;\|d\|_{2} \leq \epsilon_z\right\}
\end{equation*}

Generally, a linear approximation~\citep{goodfellow2014explaining, miyato2016adversarial} of adversarial perturbation vector $d_{adv}$ under $L_2$ norm (Fast Gradient Method) is:
\begin{equation*}
    d_{\mathrm{adv}} \approx \epsilon_z \frac{g}{\|g\|_{2}}
\end{equation*}
where $g=\nabla_{z} \log p(r|z,\theta)$, which can be efficiently computed by back-propagation in neural networks.
With such a perturbation, our maximization objective is marked as:
\begin{equation*}
\operatorname{LDS-Z}\left(\theta\right) := \sum_{z} l_{\mathrm{bat}}\left(d_{adv}, z, \theta\right)
\end{equation*}

\subsection{Objective}

Considering original objective of MIL framework mentioned in Section~\ref{sec:selective attention}, and two regularization terms at instance level (\ref{sec:IVAT}) and bag level (\ref{sec:bat}), the overall objective function of our method is:
\begin{equation*}
\mathcal{L}=J(\theta)+\beta_1 \operatorname{LDS-X}(\theta)+\beta_2 \operatorname{LDS-Z}(\theta)
\end{equation*}
where $\beta_1>0$ and $\beta_2>0$ are the weight coefficients corresponding to the modules IVAT and BAT. Module IVAT uses a hyperparameter $T_\alpha$ to decide extra-learning data ratio. Empirically, the value of $\beta_1$ is closely related to the value of parameter $T_\alpha$.

\section{Experiments}

Our experiments are designed to verify the effectiveness of MULTICAST.

\subsection{Dataset}

We evaluate our model on the widely used DSRE dataset --- NYT~\citep{mintz2009distant}, which aligns Freebase~\citep{bollacker2008freebase} entity relation with New York Times corpus. This dataset collects corpus from 2005 to 2006 as the training set, and makes corpus of 2007 as an extra test set. In detail, the training set consists of 522,611 sentences, 281,270 entity pairs and 18,252 relation facts, while the testing set contains 172,448 sentences, 96,678 entity pairs, and 1,950 relation facts. For relation labels, this dataset supports 53 different relations including NA which means no relation between an entity pair.

It is worth noting that, some previous work~\citep{wu2017adversarial,qin2018dsgan,ye2019distant} use another dataset which contains 578,288 sentences in the training set. 
In fact, that dataset is inaccurate because there is considerable training data and test data overlaps.
This bug was fixed in March 2018, and all recent work~\citep{huang2019self,shang2020noisy} since then adopt the correct dataset as the benchmark. 
In order to ensure the fairness and scientificity of our experiments, we use the original dataset release and employ the popular RE toolkit OpenNRE~\citep{han2019opennre} in our study .

\subsection{Evaluation Metrics}

Following previous literature, we conduct held-out evaluation.
Specifically, Precision-Recall curves (PR-curve) are drawn to show the trade-off between model precision and recall, the Area Under Curve (AUC) metric is used to evaluate the overall model performances, and the Precision at N (P@N) metric is also reported to consider the accuracy value for different cut-offs (default using all sentences for each entity pair while testing).

\subsection{Baseline Models}

We choose six recent methods as baseline models.

\paragraph{PCNN-ATT}
~\citep{lin2016neural} uses selective attention to reduce the weights of noisy instances.
\paragraph{PCNN-ATT+ADV}
~\citep{wu2017adversarial} adds adversarial noise to DS training data.
\paragraph{PCNN-ATT+DSGAN}
~\citep{qin2018dsgan} utilizes GANs to remove potentially inaccurate sentences from the original training dataset.
\paragraph{PCNN-ATT-RA+BAG-ATT}
~\citep{ye2019distant} uses intra-bag and inter-bag attentions to deal with the noise at sentence-level and bag-level.
\paragraph{PCNN-ATT+SELF-ATT+[CCL-CT]}
~\citep{huang2019self} integrates self-attention mechanism into the CNN structure and defines two student models for collaborative curriculum learning.
\paragraph{PCNN-ATT+DC}
~\citep{shang2020noisy} employs unsupervised deep clustering to generate reliable labels for noisy sentences.

\subsection{Overall Comparison}

\begin{table*}
\centering
\begin{tabular}{lccccc}
\hline
\textbf{Methods}          & \textbf{AUC}            & \textbf{P@100}        & \textbf{P@200}        & \textbf{P@300}        & \textbf{P@Mean}       \\ \hline
PCNN-ATT                  & 34.13                   & 73.0                  & 69.0                  & 66.0                  & 69.3                  \\
PCNN-ATT+ADV              & 34.99                   & 80.2                  & 72.1                  & 69.4                  & 73.9                  \\
PCNN-ATT-RA+BAG-ATT       & 35.03                   & 77.0                  & 75.5                  & 72.3                  & 74.9                  \\
PCNN-ATT+DSGAN            & 35.19                   & 76.2                  & 70.7                  & 68.4                  & 71.8                  \\
PCNN-ATT+SELF-ATT*        & 36.80                   & 81.1                  & 71.6                  & 70.4                  & 74.4                  \\
PCNN-ATT+SELF-ATT+CCL-CT* & 38.10                   & 82.2                  & 79.1                  & 73.1                  & 78.1                  \\ \hline
\textbf{PCNN-ATT+\our (Ours)} & \textbf{38.78$\pm$0.15} & \textbf{83.7$\pm$1.5} & \textbf{79.2$\pm$1.0} & \textbf{74.2$\pm$0.7} & \textbf{79.0$\pm$0.6} \\ \hline
\end{tabular}
\caption{\label{overall performance table}
Performances of all compared models. Models marked with * are quoted from original papers, since there are no open-source codes released.}
\vspace{-0.1cm}
\end{table*}

We summarize the model performances of our method and above-mentioned baseline models in Table \ref{overall performance table}. From the results, we can observe that: 
(1) With the help of our proposed modules (\our, \ie, IVAT+BAT), the vanilla baseline model PCNN-ATT achieves the best performance in all five metrics.
(2) Compared with the standard baseline model PCNN-ATT, \our improves the metric AUC (34.13$\to$38.78) by 13.6\% and the metric P@Mean (69.3$\to$79.0) by 14.0\%.

\begin{figure}[htbp]
\centerline{\includegraphics[scale=0.55]{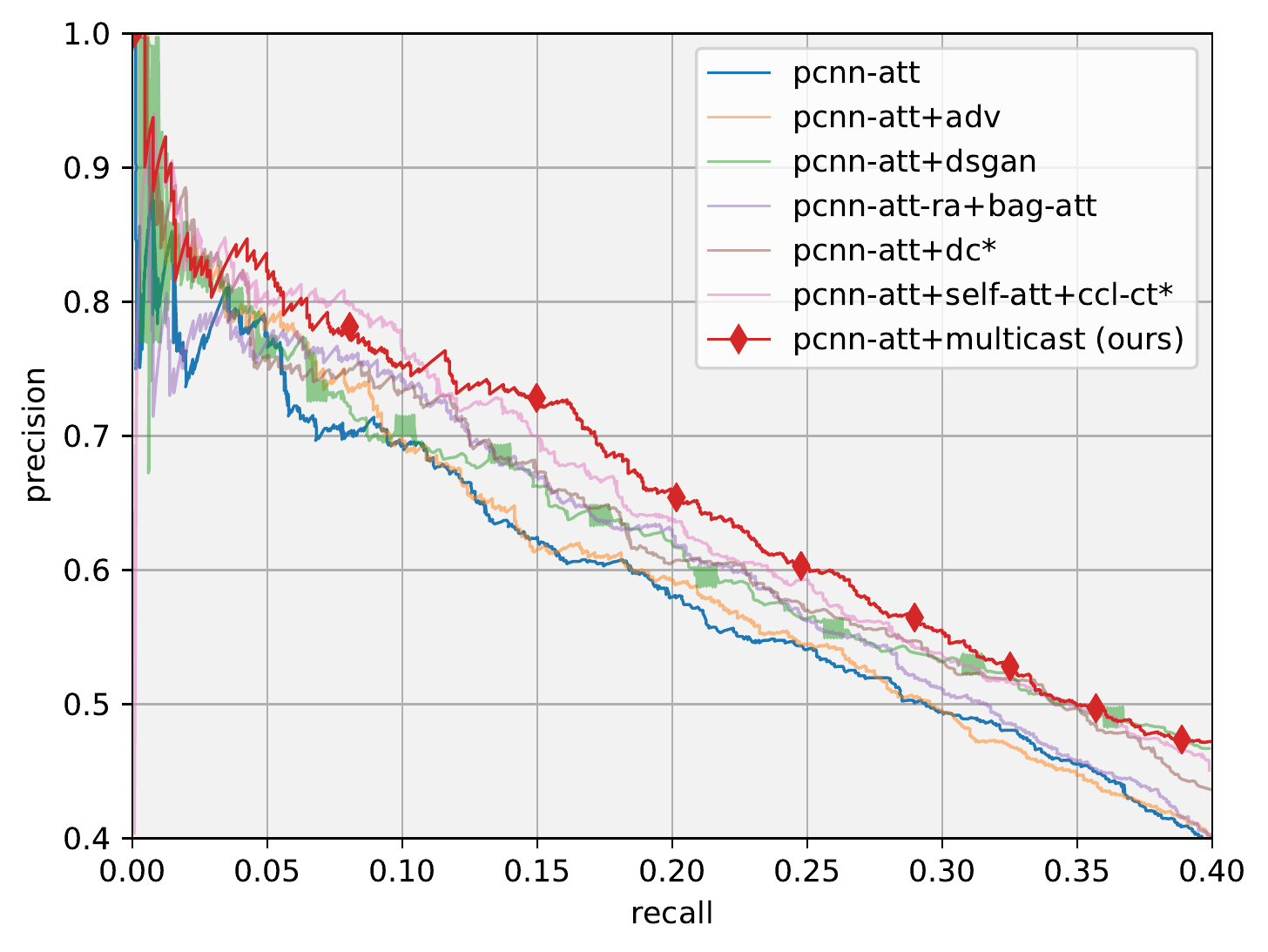}}
\caption{PR-Curve. Models with * directly quote the drawn curves from the corresponding papers.}
\label{Overall PR-Curve}
\vspace{-0.1cm}
\end{figure}

The overall PR-curve is visualized in Figure~\ref{Overall PR-Curve}. 
From the curve, we can observe that: 
(1) Compared to the PR-curve of standard baseline model PCNN-ATT, our method shifts up the curve a lot. 
(2) Our method surpasses current SOTA model in almost all ranges (except when the recall is between 0.05 and 0.10) along the curve.

\subsection{Controlled Experiment}

We identify the \emph{low data utilization} issue as the key bottleneck of DS-MIL. 
To verify that those low-score sentences are not used by the model, we remove these sentences from the training set with different thresholds (e.g. $\alpha_i < 0.1,0.2$), and use the reduced dataset to re-train PCNN-ATT models and our proposed models.

\begin{table}[htbp]
\centering
\begin{tabular}{clc}
\hline
\textbf{Dataset Size}          & \textbf{Methods} & \textbf{AUC}  \\ \hline
522611        & \small PCNN-ATT         & 34.13         \\ \cline{2-3} 
(unfiltered)  & \small +\our        & 38.93         \\ \hline
334194(-36\%) & \small PCNN-ATT         & 33.87(-0.7\%) \\ \cline{2-3} 
(filtered @ 0.1) & \small +\our        & 36.50(-6.2\%) \\ \hline
310039(-41\%) & \small PCNN-ATT         & 33.70(-1.3\%) \\ \cline{2-3} 
(filtered @ 0.2)    & \small +\our        & 36.24(-6.9\%) \\ \hline
\end{tabular}
\caption{\label{control exp}Model performances of the original dataset and reduced dataset.}
\vspace{-0.1cm}
\end{table}

We summarize model performances on the original dataset and reduced dataset in Table~\ref{control exp}. 
With the significant reduction in the amount of data, the MIL method PCNN-ATT only has subtle performance changes (i.e., yielding $\sim 1\%$ performance loss).
It verifies our intuition that MIL abandons these instances and ignores them during training. 
Besides, our method has a noticeable large performance drop (38.93$\to$36.50) after removing these training instances. 
It verifies that our proposed method effectively recycles abandoned training instances thus leading to a better data-utilization. 

\subsection{Ablation Study}

\begin{table*}
\centering
\begin{tabular}{lccccc}
\hline
\textbf{Methods}    & \textbf{AUC}          & \textbf{P@100}       & \textbf{P@200}       & \textbf{P@300}      & \textbf{P@Mean}      \\ \hline
PCNN-ATT            & 34.13                 & 73.0                 & 69.0                 & 66.0                & 69.3                 \\
+BAT                & 35.10(+0.97)          & 79.0(+6.0)           & 77.5(+8.5)           & 70.7(+4.7)          & 75.7(+6.4)           \\
+IVAT               & 37.97(+3.84)          & 81.2(+8.2)           & 77.6(+8.6)           & 73.1(+7.1)          & 77.3(+8.0)           \\
+IVAT+BAT           & \textbf{38.93}(+4.80) & \textbf{86.2}(+13.2) & \textbf{78.6}(+9.6)  & \textbf{74.1}(+8.1) & \textbf{79.6}(+10.3) \\ \hline
PCNN-ATT-RA+BAG-ATT & 35.03                 & 77.0                 & 75.5                 & 72.3                & 74.9                 \\
+IVAT*              & \textbf{38.23}(+3.20) & \textbf{87.0}(+10.0) & \textbf{82.5}(+7.0)  & \textbf{75.3}(+3.0) & \textbf{81.6}(+6.7)  \\ \hline
PCNN-ATT+DSGAN      & 35.19                 & 76.2                 & 70.7                 & 68.4                & 71.8                 \\
+BAT                & 36.24(+1.05)          & 79.2(+3.0)           & 73.1(+2.4)           & 71.8(+3.4)          & 74.7(+2.9)           \\
+IVAT               & 39.21(+4.02)          & 84.2(+8.0)           & 77.6(+6.9)           & 73.4(+5.0)          & 78.4(+6.6)           \\
+IVAT+BAT           & \textbf{40.85}(+5.66) & \textbf{86.2}(+10.0) & \textbf{81.1}(+10.4) & \textbf{74.4}(+6.0) & \textbf{80.6}(+8.8)  \\ \hline
\end{tabular}
\caption{\label{ablation table}
Ablation study with three baseline models. The model marked with * does not have bag representation and cannot be integrated with BAT (it employs a two-layer attention mechanism to get relation-aware embedding).
} 
\vspace{-0.1cm}
\end{table*}

We further conduct ablation studies to verify the effectiveness of our proposed modules respectively.

As to module IVAT, it is designed for utilizing training instances that are abandoned by the MIL framework. 
Intuitively, improvements from this module are orthogonal to attempts aiming to further improve the supervision quality. 
Thus, we add this module to three baselines and summarize their performances in Table~\ref{ablation table}.
Module IVAT brings stable and significant improvements to different baseline models in all metrics.
For the standard baseline model PCNN-ATT, with the IVAT module alone, its AUC score is already close to the current SOTA model (37.97$\sim$38.10).
For another two baseline models, module IVAT also leads to consistent performance improvements. 
For instance, with IVAT, all metrics of the method PCNN-ATT-RA+BAG-ATT have exceeded the SOTA model (e.g., its P@N score and AUC score reaches 81.6 and 38.23, while the SOTA gets 78.1 and 38.10). 

For module BAT, it aims to make full use of high-quality representations at bag level. 
As this module still relies on supervised label information (i.e., the more accurate representation is, the more model performance can be enhanced), it has fewer performance improvements than IVAT. 
Still, this module leads to consistent performance improvements on both baselines that have bag-level representations. 
Besides, we find the two modules are not exclusive (0.97+3.84$\sim$4.80, 1.05+4.02$\sim$5.66). 
More discussions are conducted in Section~\ref{sec:case study}.

\begin{figure}[htbp]
\centerline{\includegraphics[scale=0.55]{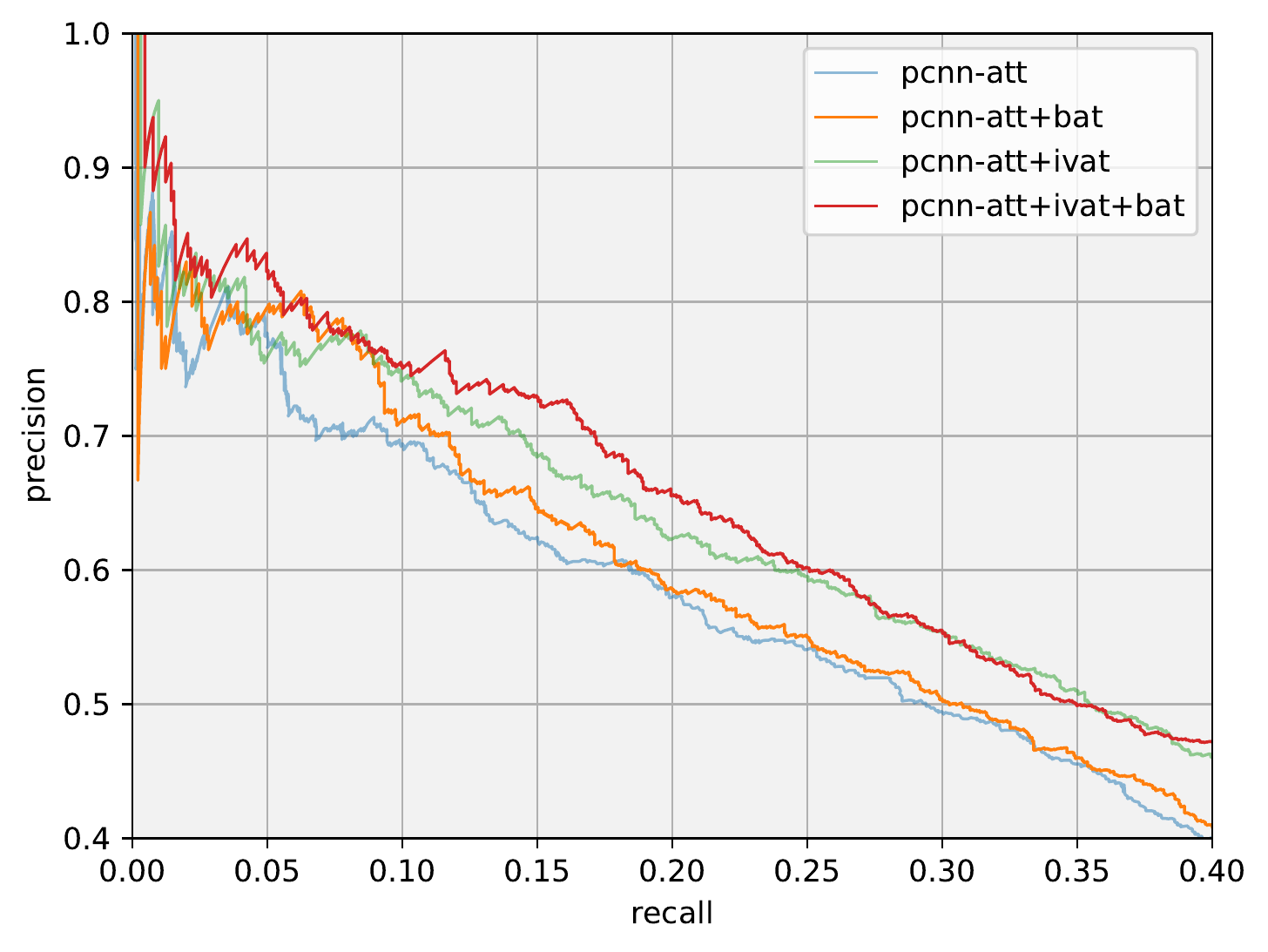}}
\caption{PR-curves of modules IVAT and BAT.}
\label{ablation curve}
\vspace{-0.1cm}
\end{figure}

To better understand the effects of two modules, we draw their PR-curves in Figure~\ref{ablation curve}. 
From the figure we can observe that: 
(1) the IVAT module significantly raises the curve of baseline model in all ranges. 
(2) the BAT module has a larger benefit with a higher precision score. 
This observation further verifies our intuitions.

\subsection{Discussion About AT and VAT}

Our method \our leverage two strategies to coordinate AT and VAT: (1) instead of adding AT/VAT to all instances, MIL attention signals are leveraged to recognize the proper subset to apply these techniques. 
(2) instead of applying both AT and VAT at both levels, we only apply AT at the bag level and VAT at the instance level.

\begin{table}[htbp]
\centering
\begin{tabular}{lcc}
\hline
\textbf{Methods}              & \textbf{Level}  & \textbf{AUC}   \\ \hline
PCNN-ATT                      & -               & 34.13          \\ \hline
\multirow{2}{*}{PCNN-ATT+AT}  & all instances   & 34.99          \\ \cline{2-3} 
                              & bag features    & \textbf{35.10} \\ \hline
\multirow{2}{*}{PCNN-ATT+VAT} & all instances   & 37.35          \\ \cline{2-3} 
                              & noisy instances & \textbf{37.97} \\ \hline
\end{tabular}
\caption{Discussion of different level selection ways.}
\label{ablation strategy table}
\vspace{-0.1cm}
\end{table}

\paragraph{Effectiveness of Level Selection}

Classical methods apply AT~\citep{wu2017adversarial}/VAT to all instances without any selection. 
In contrast, \our applies AT and VAT at different levels.
To verify the effectiveness of this strategy, we conduct comparison to the conventional methods and summarize the results in Table~\ref{ablation strategy table}.
The gap between adding AT to all instances and adding AT to bag features are marginal. 
Intuitively, these two methods are very similar to each other, while adding AT at the bag-level is faster (no need to conduct back-propagate to the embedding layer). 
On the other hand, adding VAT to all more instances (which would also be slower) performs worse than only adding VAT to abandoned instances. 
It verifies that the context information of high-quality instances are already utilized by the training algorithm, and there is no need to apply VAT on these instances. 

\paragraph{Effectiveness of Collaboration}

\begin{table}[htbp]
\centering
\begin{tabular}{lc}
\hline
\textbf{Methods}                                                        & \textbf{AUC}   \\ \hline
PCNN-ATT                                                                & 34.13          \\
+Instance-Level \phantom{VAT}\llap{AT}+Bag-Level VAT                    & 32.34          \\
+Instance-Level \phantom{VAT}\llap{AT}+Bag-Level \phantom{VAT}\llap{AT} & 34.16          \\
+Instance-Level VAT+Bag-Level VAT                                       & 36.36          \\
+Instance-Level VAT+Bag-Level \phantom{VAT}\llap{AT}                    & \textbf{38.93} \\ \hline
\end{tabular}
\caption{Discussion of different collaboration ways.}
\label{ablation combine table}
\vspace{-0.1cm}
\end{table}
\begin{table*}[htbp]
\centering
\begin{tabular}{lcccc}
\hline
\multicolumn{5}{c}{KB Fact: (\textit{lebron james} \textsl{lived\_in} \textit{akron})\quad Bag Label: \textit{/people/person/place\_lived}}                                                                                                 \\ \hline
\multirow{2}{*}{Sentences}                                                                                                                                       & \multicolumn{2}{c}{Attention Score} & \multicolumn{2}{c}{Sentence Label} \\ \cline{2-5} 
                                                                                                                                                                 & w/o BAT           & w/ BAT          & w/o IVAT        & w/ IVAT          \\ \hline
\begin{tabular}[c]{@{}l@{}}an estimated 40,000 ohio state fans came to town, \\ including the \textbf{akron} native \textbf{lebron james}, ...\end{tabular}      & 0.59              & 0.71            & lived\_in       & lived\_in        \\ \hline
\begin{tabular}[c]{@{}l@{}}bynum is not another \textbf{lebron james}, the high school\\ phenomenon from \textbf{akron}, ohio, who was the top ...\end{tabular}  & 0.19              & 0.13            & NA              & borned\_in       \\ \hline
\begin{tabular}[c]{@{}l@{}}\textbf{lebron james} and his friends used to drive from \\ \textbf{akron}, ohio, fill a few of the empty aquamarine ...\end{tabular} & 0.22              & 0.16            & lived\_in       & NA               \\ \hline
\end{tabular}
\caption{\label{case table}Case study of how modules IVAT and BAT work.}
\vspace{-0.1cm}
\end{table*}
After clarifying the choice of level selection, we proceed to consider the cooperation strategy between AT and VAT (results are summarized in Table~\ref{ablation combine table}): 
(1) For instance-level noisy data, AT may amplify the effects of wrong labels and results in severe \emph{confirmation bias problem}~\citep{tarvainen1703weight}, which makes the model converge too fast and learn nothing extra (34.13$\sim$34.16). 
(2) For bag-level high-quality features, VAT may weaken the original supervised information provided by MIL framework and complicates model training (38.93$\to$36.36, 34.16$\to$32.34). 
Comparing Table~\ref{ablation strategy table} and Table~\ref{ablation combine table}, 
Instance-Level AT and Bag-Level VAT actually has a negative impact on the model performance (35.10$\to$34.16, 37.35$\to$36.36).

\subsection{Case Study and Visualization}
\label{sec:case study}

In order to better understand
how BAT and IVAT work, we conduct case studies and visualization. 


In our method \our, both IVAT and BAT improve model performance, but the ways they work are entirely different.
We select a typical bag in the training set to illustrate their roles respectively:
(1) For the bag (see Table~\ref{case table}) with KB fact (\textit{lebron james} \textsl{lived\_in} \textit{akron}), it consists of three different sentences. 
Module IVAT pays its attention to these low-score (0.19, 0.22) sentences. 
With the help of the IVAT module, these sentences are allowed to rethink their probability distributions without considering their noisy labels. 
For example, although the $3rd$ sentence (\textit{lebron james and his friends used to drive from akron ...}) mentions the entity pair (\textit{lebron}, \textit{james}), it actually fails to express the relation \textsl{live\_in}. 
With the help of IVAT, this instance succeeds to realize the error and find its true label to be NA.
(2) Meanwhile, the BAT module focuses on accurate bag features formed by high-quality instances. In this bag, the final representation is mainly composed of the $1st$ sentence (\textit{... including the akron native lebron james}), which is representative enough to express current bag label \textsl{lived\_in}.
After the adversarial enhancement at bag level, the model is more confident in the high-quality instance with higher attention score (the representation of the $1st$ instance is near to the bag-level representation).

\begin{figure}[t]
\centerline{\includegraphics[width=\linewidth]{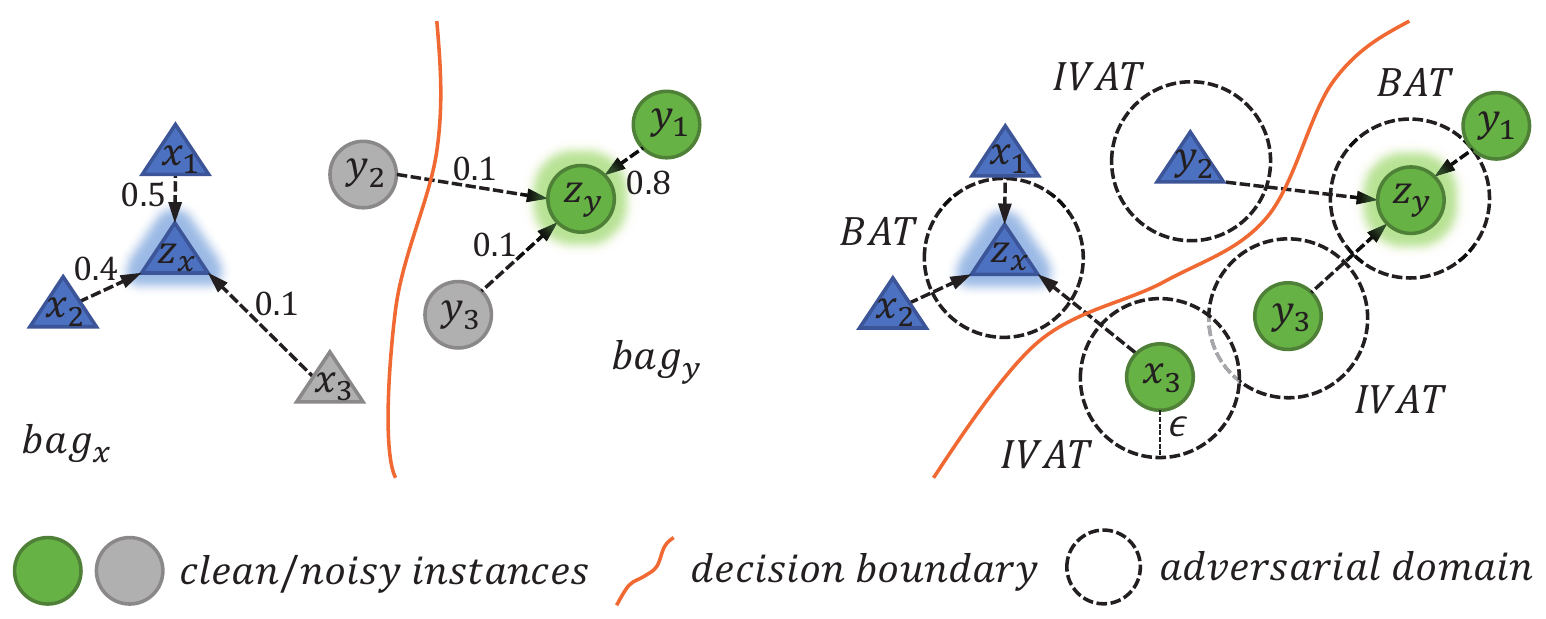}}
\caption{Effect diagram of modules IVAT and BAT.}
\label{IVAT+BAT}
\vspace{-0.1cm}
\end{figure}

Moreover, we draw a diagram to better illustrate their mechanisms.
On the left side of Figure~\ref{IVAT+BAT}, original DS-MIL only uses bag features $z_x$ and $z_y$ for training and obtains a decision boundary without considering noisy instances like $y_2$. 
Thus, the resulting model may not be trained with abundant instances and have issues like shifted label distribution~\cite{ye2019looking}. 
On the right side, IVAT helps instances $x_3$ and $y_2$ find their correct labels.
It works with BAT to smooth model outputs in their respective adversarial domains, which prompts the model to generate a better classification boundary.

From the above diagram we can also see that, IVAT acts on those noisy instances ($x_3,y_2,y_3$), which are far away from the targets of module BAT --- bag features $z_x,z_y$. 
Therefore, the adversarial domains of modules BAT and IVAT only have limited overlap, which provides an intuitive explanation for why the effects of two modules are orthogonal (see Table~\ref{ablation table}, 0.97+3.84$\sim$4.80, 1.05+4.02$\sim$5.66).

\section{Conclusion}

In this paper, we propose Multi-Instance Collaborative Adversarial Training (MULTICAST) to alleviate the problem of low data utilization under MIL framework.
Experiments have shown the effectiveness of our method with stable and significant improvements over several different baseline models, including current SOTA systems.

\section*{Acknowledgments}

This work has been supported in part by National Key Research and Development Program of China (2018AAA010010), Zhejiang NSF (LR21F020004), Zhejiang University iFLYTEK Joint Research Center, Funds from City Cloud Technology (China) Co. Ltd., Zhejiang University-Tongdun Technology Joint Laboratory of Artificial Intelligence, Chinese Knowledge Center of Engineering Science and Technology (CKCEST), Hikvision-Zhejiang University Joint Research Center, the Fundamental Research Funds for the Central Universities, and Engineering Research Center of Digital Library, Ministry of Education. 

\bibliography{anthology,emnlp2020}

\begin{thebibliography}{19}
\expandafter\ifx\csname natexlab\endcsname\relax\def\natexlab#1{#1}\fi

\bibitem[{Bollacker et~al.(2008)Bollacker, Evans, Paritosh, Sturge, and
  Taylor}]{bollacker2008freebase}
Kurt Bollacker, Colin Evans, Praveen Paritosh, Tim Sturge, and Jamie Taylor.
  2008.
\newblock Freebase: a collaboratively created graph database for structuring
  human knowledge.
\newblock In \emph{Proceedings of the 2008 ACM SIGMOD international conference
  on Management of data}, pages 1247--1250.

\bibitem[{Goodfellow et~al.(2015)Goodfellow, Shlens, and
  Szegedy}]{goodfellow2014explaining}
Ian~J. Goodfellow, Jonathon Shlens, and Christian Szegedy. 2015.
\newblock \href {http://arxiv.org/abs/1412.6572} {Explaining and harnessing
  adversarial examples}.
\newblock In \emph{3rd International Conference on Learning Representations,
  {ICLR} 2015, San Diego, CA, USA, May 7-9, 2015, Conference Track
  Proceedings}.

\bibitem[{Han et~al.(2019)Han, Gao, Yao, Ye, Liu, and Sun}]{han2019opennre}
Xu~Han, Tianyu Gao, Yuan Yao, Deming Ye, Zhiyuan Liu, and Maosong Sun. 2019.
\newblock \href {https://doi.org/10.18653/v1/D19-3029} {{O}pen{NRE}: An open
  and extensible toolkit for neural relation extraction}.
\newblock In \emph{Proceedings of the 2019 Conference on Empirical Methods in
  Natural Language Processing and the 9th International Joint Conference on
  Natural Language Processing (EMNLP-IJCNLP): System Demonstrations}, pages
  169--174, Hong Kong, China. Association for Computational Linguistics.

\bibitem[{Hoffmann et~al.(2011)Hoffmann, Zhang, Ling, Zettlemoyer, and
  Weld}]{hoffmann2011knowledge}
Raphael Hoffmann, Congle Zhang, Xiao Ling, Luke Zettlemoyer, and Daniel~S.
  Weld. 2011.
\newblock \href {https://www.aclweb.org/anthology/P11-1055} {Knowledge-based
  weak supervision for information extraction of overlapping relations}.
\newblock In \emph{Proceedings of the 49th Annual Meeting of the Association
  for Computational Linguistics: Human Language Technologies}, pages 541--550,
  Portland, Oregon, USA. Association for Computational Linguistics.

\bibitem[{Huang and Du(2019)}]{huang2019self}
Yuyun Huang and Jinhua Du. 2019.
\newblock \href {https://doi.org/10.18653/v1/D19-1037} {Self-attention enhanced
  {CNN}s and collaborative curriculum learning for distantly supervised
  relation extraction}.
\newblock In \emph{Proceedings of the 2019 Conference on Empirical Methods in
  Natural Language Processing and the 9th International Joint Conference on
  Natural Language Processing (EMNLP-IJCNLP)}, pages 389--398, Hong Kong,
  China. Association for Computational Linguistics.

\bibitem[{Lin et~al.(2016)Lin, Shen, Liu, Luan, and Sun}]{lin2016neural}
Yankai Lin, Shiqi Shen, Zhiyuan Liu, Huanbo Luan, and Maosong Sun. 2016.
\newblock \href {https://doi.org/10.18653/v1/P16-1200} {Neural relation
  extraction with selective attention over instances}.
\newblock In \emph{Proceedings of the 54th Annual Meeting of the Association
  for Computational Linguistics (Volume 1: Long Papers)}, pages 2124--2133,
  Berlin, Germany. Association for Computational Linguistics.

\bibitem[{Mintz et~al.(2009)Mintz, Bills, Snow, and
  Jurafsky}]{mintz2009distant}
Mike Mintz, Steven Bills, Rion Snow, and Daniel Jurafsky. 2009.
\newblock \href {https://www.aclweb.org/anthology/P09-1113} {Distant
  supervision for relation extraction without labeled data}.
\newblock In \emph{Proceedings of the Joint Conference of the 47th Annual
  Meeting of the {ACL} and the 4th International Joint Conference on Natural
  Language Processing of the {AFNLP}}, pages 1003--1011, Suntec, Singapore.
  Association for Computational Linguistics.

\bibitem[{Miyato et~al.(2017)Miyato, Dai, and
  Goodfellow}]{miyato2016adversarial}
Takeru Miyato, Andrew~M. Dai, and Ian~J. Goodfellow. 2017.
\newblock \href {https://openreview.net/forum?id=r1X3g2\_xl} {Adversarial
  training methods for semi-supervised text classification}.
\newblock In \emph{5th International Conference on Learning Representations,
  {ICLR} 2017, Toulon, France, April 24-26, 2017, Conference Track
  Proceedings}. OpenReview.net.

\bibitem[{Miyato et~al.(2018)Miyato, Maeda, Koyama, and
  Ishii}]{miyato2018virtual}
Takeru Miyato, Shin-ichi Maeda, Masanori Koyama, and Shin Ishii. 2018.
\newblock Virtual adversarial training: a regularization method for supervised
  and semi-supervised learning.
\newblock \emph{IEEE transactions on pattern analysis and machine
  intelligence}, 41(8):1979--1993.

\bibitem[{Qin et~al.(2018)Qin, Xu, and Wang}]{qin2018dsgan}
Pengda Qin, Weiran Xu, and William~Yang Wang. 2018.
\newblock \href {https://doi.org/10.18653/v1/P18-1046} {{DSGAN}: Generative
  adversarial training for distant supervision relation extraction}.
\newblock In \emph{Proceedings of the 56th Annual Meeting of the Association
  for Computational Linguistics (Volume 1: Long Papers)}, pages 496--505,
  Melbourne, Australia. Association for Computational Linguistics.

\bibitem[{Riedel et~al.(2010)Riedel, Yao, and McCallum}]{riedel2010modeling}
Sebastian Riedel, Limin Yao, and Andrew McCallum. 2010.
\newblock Modeling relations and their mentions without labeled text.
\newblock In \emph{Joint European Conference on Machine Learning and Knowledge
  Discovery in Databases}, pages 148--163. Springer.

\bibitem[{Shang et~al.(2020)Shang, Huang, Mao, Sun, and Wei}]{shang2020noisy}
Yuming Shang, He~Yan Huang, Xianling Mao, Xin Sun, and Wei Wei. 2020.
\newblock Are noisy sentences useless for distant supervised relation
  extraction?
\newblock In \emph{AAAI}, pages 8799--8806.

\bibitem[{Tarvainen and Valpola(2017)}]{tarvainen1703weight}
Antti Tarvainen and Harri Valpola. 2017.
\newblock Weight-averaged consistency targets improve semi-supervised deep
  learning results.
\newblock \emph{CoRR abs/1703.01780. arXiv}, 1703:01780.

\bibitem[{Vaswani et~al.(2017)Vaswani, Shazeer, Parmar, Uszkoreit, Jones,
  Gomez, Kaiser, and Polosukhin}]{vaswani2017attention}
Ashish Vaswani, Noam Shazeer, Niki Parmar, Jakob Uszkoreit, Llion Jones,
  Aidan~N. Gomez, Lukasz Kaiser, and Illia Polosukhin. 2017.
\newblock \href
  {https://proceedings.neurips.cc/paper/2017/hash/3f5ee243547dee91fbd053c1c4a845aa-Abstract.html}
  {Attention is all you need}.
\newblock In \emph{Advances in Neural Information Processing Systems 30: Annual
  Conference on Neural Information Processing Systems 2017, December 4-9, 2017,
  Long Beach, CA, {USA}}, pages 5998--6008.

\bibitem[{Wu et~al.(2017)Wu, Bamman, and Russell}]{wu2017adversarial}
Yi~Wu, David Bamman, and Stuart Russell. 2017.
\newblock \href {https://doi.org/10.18653/v1/D17-1187} {Adversarial training
  for relation extraction}.
\newblock In \emph{Proceedings of the 2017 Conference on Empirical Methods in
  Natural Language Processing}, pages 1778--1783, Copenhagen, Denmark.
  Association for Computational Linguistics.

\bibitem[{Ye et~al.(2019)Ye, Liu, Zhang, and Ren}]{ye2019looking}
Qinyuan Ye, Liyuan Liu, Maosen Zhang, and Xiang Ren. 2019.
\newblock \href {https://doi.org/10.18653/v1/D19-1397} {Looking beyond label
  noise: Shifted label distribution matters in distantly supervised relation
  extraction}.
\newblock In \emph{Proceedings of the 2019 Conference on Empirical Methods in
  Natural Language Processing and the 9th International Joint Conference on
  Natural Language Processing (EMNLP-IJCNLP)}, pages 3841--3850, Hong Kong,
  China. Association for Computational Linguistics.

\bibitem[{Ye and Ling(2019)}]{ye2019distant}
Zhi-Xiu Ye and Zhen-Hua Ling. 2019.
\newblock \href {https://doi.org/10.18653/v1/N19-1288} {Distant supervision
  relation extraction with intra-bag and inter-bag attentions}.
\newblock In \emph{Proceedings of the 2019 Conference of the North {A}merican
  Chapter of the Association for Computational Linguistics: Human Language
  Technologies, Volume 1 (Long and Short Papers)}, pages 2810--2819,
  Minneapolis, Minnesota. Association for Computational Linguistics.

\bibitem[{Yuan et~al.(2019)Yuan, Liu, Tang, Zhang, Zhuang, Pu, Wu, and
  Ren}]{yuan2019cross}
Yujin Yuan, Liyuan Liu, Siliang Tang, Zhongfei Zhang, Yueting Zhuang, Shiliang
  Pu, Fei Wu, and Xiang Ren. 2019.
\newblock Cross-relation cross-bag attention for distantly-supervised relation
  extraction.
\newblock In \emph{Proceedings of the AAAI Conference on Artificial
  Intelligence}, volume~33, pages 419--426.

\bibitem[{Zeng et~al.(2015)Zeng, Liu, Chen, and Zhao}]{zeng2015distant}
Daojian Zeng, Kang Liu, Yubo Chen, and Jun Zhao. 2015.
\newblock \href {https://doi.org/10.18653/v1/D15-1203} {Distant supervision for
  relation extraction via piecewise convolutional neural networks}.
\newblock In \emph{Proceedings of the 2015 Conference on Empirical Methods in
  Natural Language Processing}, pages 1753--1762, Lisbon, Portugal. Association
  for Computational Linguistics.

\end{thebibliography}
\bibliographystyle{acl_natbib}

\end{document}